\documentclass[sigconf]{acmart}
\usepackage{xcolor}
\usepackage{booktabs} 
\usepackage{array}    
\usepackage{multirow, multicol}
\usepackage{algorithm}
\usepackage{algorithmic}
\usepackage{float}

\AtBeginDocument{%
  }

\copyrightyear{2026}
\acmYear{2026}
\setcopyright{cc}
\setcctype{by}
\acmConference[MM '26]{Proceedings of the 34th ACM International Conference on Multimedia}{November 10--14, 2026}{Rio de Janeiro, Brazil}
\acmBooktitle{Proceedings of the 34th ACM International Conference on Multimedia (MM '26), November 10--14, 2026, Rio de Janeiro, Brazil}
\acmDOI{10.1145/3767308.3836202}
\acmISBN{979-8-4007-2213-4/2026/11}

\begin{document}

\title{VTO: Visual Tool Orchestration for Video Anomaly Detection}



\author{Rui Wang}
\authornote{Rui Wang and Yeteng Wu contributed equally to this work.}
\orcid{0000-0001-5919-0847}
\affiliation{%
  \department{State Key Laboratory of Networking and Switching Technology}
  \institution{Beijing University of Posts and Telecommunications}
  \city{Beijing}
  \country{China}
}
\email{wr@bupt.edu.cn}

\author{Yeteng Wu}
\authornotemark[1]
\orcid{0009-0000-3680-6178}
\affiliation{%
  \department{State Key Laboratory of Networking and Switching Technology}
  \institution{Beijing University of Posts and Telecommunications}
  \city{Beijing}
  \country{China}
}
\email{wuyeteng@bupt.edu.cn}

\author{Xianlin Zhang}
\orcid{0000-0003-3905-2062}
\affiliation{%
  \department{School of Digital Media \& Design Art}
  \institution{Beijing University of Posts and Telecommunications}
  \city{Beijing}
  \country{China}
}
\email{zxlin@bupt.edu.cn}

\author{Mengshi Qi}
\authornote{Corresponding author: Mengshi Qi.}
\orcid{0000-0002-6955-6635}
\affiliation{%
  \department{State Key Laboratory of Networking and Switching Technology}
  \institution{Beijing University of Posts and Telecommunications}
  \city{Beijing}
  \country{China}
}
\email{qms@bupt.edu.cn}

\renewcommand{\shortauthors}{Wang et al.}

\begin{abstract}

Video anomaly detection (VAD) is a critical yet challenging task due to the complex and diverse nature of real-world scenarios. Traditional deep learning approaches are fundamentally limited by poor generalization across diverse scenarios. While multimodal agents offer a promising tool-learning paradigm for VAD, current systems relying on supervised fine-tuning struggle with complex orchestration, and standard reinforcement learning often causes premature termination due to coarse-grained outcome rewards. To address these challenges, we propose VTO, a process-supervised reinforcement learning framework. Moving beyond static tool usage, VTO enables the agent to dynamically explore and interact with the environment. Specifically, we introduce a foundation model-driven cognitive evaluator to provide context-aware semantic feedback, which is seamlessly integrated into a Process-Supervised Cognitive Alignment that delivers fine-grained, step-wise supervision. By explicitly penalizing logical truncation and rewarding complete causal chains, the agent optimizes its multi-step reasoning policy for interrelated tool orchestration. To support our proposed framework, we meticulously crafted VAD-Tool, a hierarchical visual tool set comprising 12 specialized vision tools spanning from entity tracking to high-stakes hazard detection, and established the corresponding benchmark for rigorous multi-step reasoning evaluation. Extensive experiments on VAD-Tool demonstrate that VTO significantly outperforms baselines, achieving up to a 10.2\% absolute accuracy improvement in tool scheduling. Code and data are available at \url{https://github.com/MICLAB-BUPT/VTO}.
\end{abstract}



\maketitle

\section{Introduction}
\begin{figure}[!t]
	\centering
    \includegraphics[width=0.99\columnwidth]{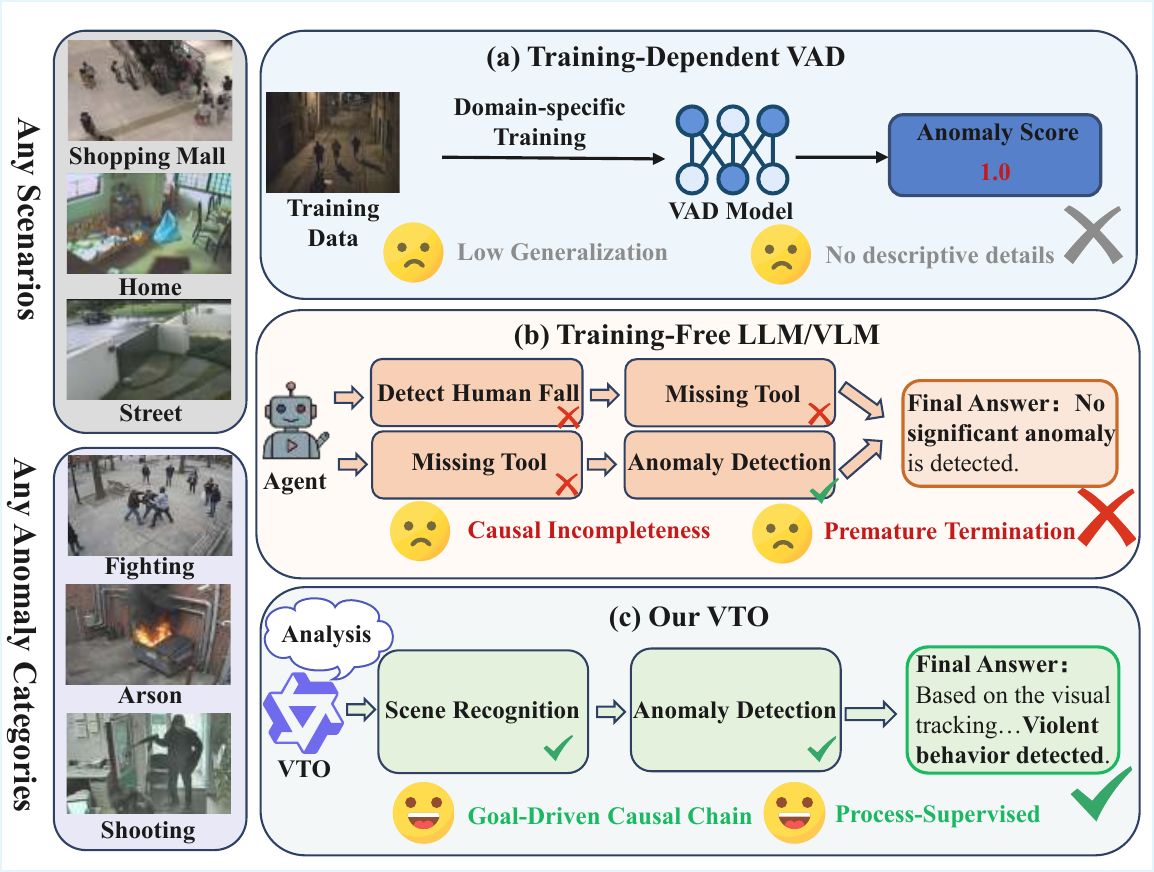}
    \caption{Comparison of VAD paradigms. (a) Traditional VAD yields low generalization. (b) Training-Free LLM/VLM agents suffer from premature termination and causal incompleteness. (c) Our VTO autonomously constructs goal-driven causal chains through process-supervised cognitive alignment, enabling accurate multi-step reasoning across diverse scenarios.}
    \label{fig: teaser}
\end{figure}

Video Anomaly Detection (VAD) is a fundamental yet challenging task in computer vision, playing a pivotal role in applications ranging from public safety to intelligent urban surveillance \cite{sultani2018real, review_vad_2025}. In real-world environments, anomaly events are rarely isolated incidents; they are inherently complex, diverse, and heavily dependent on specific physical contexts.
For instance, a local altercation can rapidly escalate into a weaponized assault, eventually triggering crowd panic and stampedes. Deciphering such composite events requires a comprehensive analysis combining multiple perceptual capabilities, such as recognizing human actions, tracking specific entities, and assessing crowd density \cite{hycovad_arxiv2025, lv2024video}. 
However, as conceptualized in Figure~\ref{fig: teaser}(a), traditional VAD paradigms primarily formulate this problem as a close-set classification or anomaly scoring task. While these methods can mathematically identify pattern deviations or predefined anomaly categories, they are fundamentally limited by poor generalization across open-world scenarios.
These severe limitations highlight the urgent need for a more generalizable and interactive anomaly reasoning paradigm capable of verbally reasoning about what the anomaly is and how it unfolds.

Recently, the community has rapidly pivoted towards leveraging Large Language Models (LLMs) and Vision-Language Models (VLMs) for explainable VAD. Pioneering works have explored training-free paradigms \cite{wu2024harnessing}, rule-based reasoning \cite{follow_rules_eccv2024}, hybrid SSL-LLM architectures \cite{hycovad_arxiv2025}, and verbalized learning via VLM dialogues \cite{vera_cvpr2025}. Furthermore, early agentic systems like PANDA \cite{PANDA_NeurIPS2025} have attempted to invoke external tools for anomaly analysis. However, these paradigms predominantly rely on Supervised Fine-Tuning (SFT) or heuristic behavior cloning, treating tools as isolated problem-solving capsules. Because they depend entirely on imitation, they struggle to dynamically orchestrate distinct visual expert models to collaboratively decipher the intricate causal dependencies inherent in complex anomaly scenes. To transcend these limitations, one might intuitively apply standard Reinforcement Learning (RL) for dynamic policy optimization. Yet, directly applying outcome-based RL introduces a critical flaw in VAD scenarios: premature termination. Constrained by coarse-grained, delayed rewards, standard RL agents tend to halt tool execution immediately after detecting a primary, isolated anomaly. Without fine-grained, step-wise supervision, they are implicitly encouraged to conclude their analysis early, thereby ignoring cascading secondary hazards and failing to complete the necessary multi-step causal chain.

To address these limitations, we redefine Video Anomaly Detection as a complex, multi-step cognitive tool orchestration problem. We propose \textbf{VTO}, a process-supervised reinforcement learning framework explicitly designed for rigorous anomaly reasoning. Instead of solving isolated sub-tasks via static mapping, our agent actively interacts with video environments by iteratively generating reasoning thoughts, invoking specific visual tools, and updating its state based on multi-modal observations. 
To break the performance upper bound of static behavior cloning and standard outcome-based RL, VTO introduces Process-Supervised Cognitive Alignments. Rather than relying on a delayed, generic final reward, this alignment mechanism provides fine-grained, step-wise supervision to strictly synchronize the agent's reasoning trajectory with expert logic. It dynamically aligns each reasoning step by synergizing rule-based objective metrics (\emph{e.g.}, exact tool match) with foundation model-driven evaluations (\emph{e.g.}, logicality and causal completeness). Driven by these dense signals, the agent optimizes its policy via Group Relative Policy Optimization (GRPO). VTO effectively overcomes premature termination, successfully mastering the detection of cascading secondary hazards before concluding its safety assessment.

To support and rigorously evaluate this framework, we establish a comprehensive infrastructure and benchmark named \textbf{VAD-Tool}. First, VAD-Tool incorporates a hierarchical visual tool set integrating 12 distinct expert models that span from spatiotemporal entity tracking and counting to high-stakes hazard detection. Second, to overcome the limitations of traditional outcome-driven datasets \cite{msad_neurips2024}, VAD-Tool provides a rich collection of interrelated anomaly queries. This establishes a rigorous testing ground specifically designed to evaluate the multi-step reasoning and dynamic tool orchestration capabilities of MLLM agents in complex, cascading hazard scenarios.

The main contributions of our work are summarized as follows:
\begin{itemize}
    \item We introduce \textbf{VTO}, a dynamic multimodal agent framework powered by process-supervised reinforcement learning. It addresses complex cognitive tool orchestration and explicitly overcomes the premature termination problem in  Video Anomaly Detection.
    \item We construct \textbf{VAD-Tool}, a comprehensive benchmark that integrates 12 specialized vision models and serves as a rigorous evaluation platform tailored for single-step and multi-step interrelated anomaly reasoning.
    \item Extensive quantitative experiments demonstrate that VTO achieves substantial accuracy improvements over static SFT baselines and also remarkably surpasses the performance of 72B-parameter models.
\end{itemize}

\section{Related Work}

\noindent \textbf{Video Anomaly Detection.}
Conventional Video Anomaly Detection (VAD) primarily formulates the problem as a dedicated, single-task objective. These approaches traditionally focus on modeling spatial-temporal patterns to detect and localize specific abnormal events within video sequences \cite{hasan2016learning, sultani2018real, gong2019memorizing, msad_neurips2024,qi2025dc,qi2025robust,qi2021semantics,qi2026explainable}. 
Recently, the community has pivoted towards leveraging Large Language Models (LLMs) and Vision-Language Models (VLMs) to generate textual anomaly descriptions. Pioneering works have explored training-free paradigms~\cite{wu2024harnessing}, rule-based reasoning~\cite{follow_rules_eccv2024}, and verbalized learning via VLM dialogues~\cite{vera_cvpr2025}.  PANDA \cite{PANDA_NeurIPS2025} attempts to invoke external tools for VAD.
However, these existing multimodal and early agentic systems predominantly operate under static paradigms, either acting as rigid end-to-end mappers or treating tools as isolated problem-solving capsules. Breaking away from these imitation-based limitations, VTO introduces Process-Supervised Cognitive Alignment to dynamically orchestrate real-world visual expert models, ensuring the construction of observable, multi-step causal chains without premature termination.

\noindent \textbf{Tool Learning.}
Tool-augmented agents offer a promising paradigm to address this attribution gap by delegating perceptual sub-tasks to specialized models~\cite{gao2025multi,zhu2026vtc,xu2024advancing_inference,iclr2026_empowering_tool,iclr2025_toolplanner,Lv_2026_CVPR, qi2025action,qi2026synergistic}. 
Early tool-augmented agents predominantly relied on static Supervised Fine-Tuning (SFT). However, as highlighted by retrospective studies \cite{xu2024advancing_inference}, relying solely on SFT constrains agents from dynamically adapting their reasoning paths, halting execution immediately after identifying a primary anomaly and fatally ignoring cascading secondary hazards \cite{ma2024agentboard}. To transcend these rigid boundaries, recent paradigms have rapidly shifted towards dynamic task planning \cite{iclr2025_toolplanner} and training tool-augmented LLMs via pure Reinforcement Learning (RL) \cite{emnlp2025_toolzero}. Furthermore, the latest state-of-the-art methods advocate for assigning fine-grained rewards to individual tool calls \cite{iclr2026_empowering_tool} to alleviate gradient conflicts during complex multi-step execution. 
Despite algorithmic leaps in structured domains, applying these dynamic reward mechanisms to real-world physical environments remains a formidable challenge. Our work bridges this gap by dynamically orchestrating real-world VAD tools, empowering agents to tackle unpredictable, cascading hazards without premature termination.

\noindent \textbf{Reinforcement Learning for Agents.}
To break free from the rigid constraints of static Supervised Fine-Tuning (SFT), recent approaches leverage reinforcement learning (RL) to dynamically optimize multi-step reasoning policies through continuous environmental feedback~\cite{shinn2023reflexion,zeng2024agenttuning,se_agent_2024,yuan2025evoagent,wang2025pitn,ye2025safedriverag,deng2025global,lv2025t2sg,lv2024sgformer}. 
Recent literature has extensively explored policy refinement via language feedback \cite{self_language_feedback}, multi-turn reinforcement learning \cite{ragen_2024}, and trajectory optimization in multi-step reasoning \cite{yuan2025evoagent}. These RL-driven architectures have demonstrated remarkable success in iterative tasks, such as automated agentic workflows for code generation \cite{sew_code_2024}.
While successful in structured domains, applying standard RL to the cascading complexities of video anomaly detection often leads to premature termination. Bridging this gap, VTO pioneers process-supervised RL in VAD, explicitly overcoming this limitation to dynamically refine multi-step tool orchestration.

\section{VAD-Tool: A Benchmark for Video Anomaly Detection.}

\subsection{Problem Formulation}
\begin{figure}[h]
    \centering
    \includegraphics[width=0.99\columnwidth]{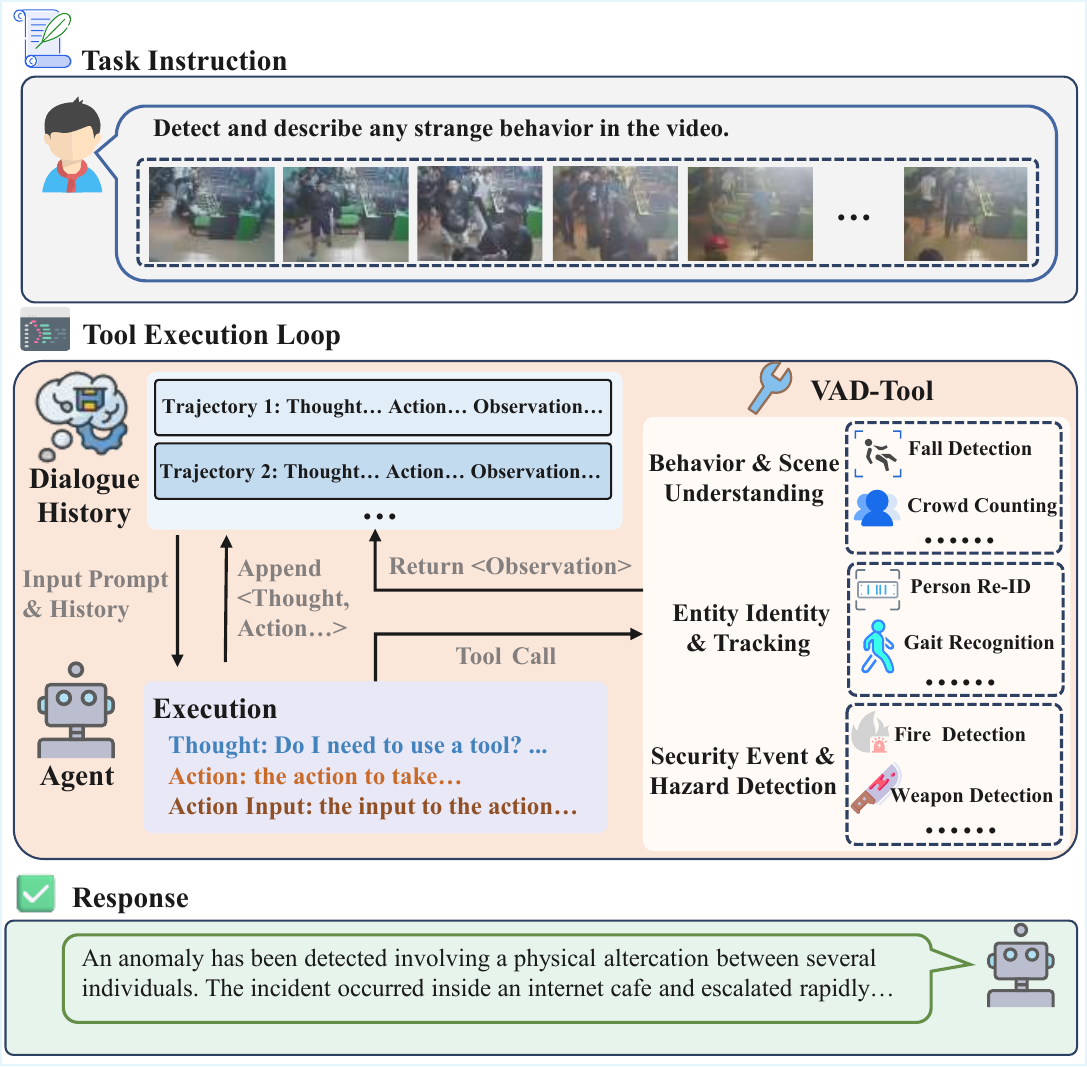}
   \caption{The inference pipeline of the proposed framework. The agent processes user queries and videos through a dynamic tool execution loop, orchestrating VAD-Tool experts to generate an explainable anomaly report.}
    \label{fig: framework}
\end{figure}

For the VTO agent task, a hierarchical visual toolset is utilized to analyze the physical semantics of a specific, user-provided visual scene. In this paradigm, the input video or image directly serves as the dynamic physical environment, while the specialized visual models constitute the agent's action space. Formally, as depicted in our overall framework in Figure~\ref{fig: framework}, given a multi-modal user query consisting of a natural language instruction $q_i$ and an accompanying visual input $v_i$, alongside a predefined visual toolset $\mathcal{T} = \{T_1, T_2, \ldots, T_m\}$, the task involves iteratively generating a sequence of reasoning actions $A_{i,t}$ to interact with $v_i$ and obtain multimodal observations $o_{i,t}$. This process yields a reasoning trajectory $C_{i,L} = \{(A_{i,1}, o_{i,1}), \dots, (A_{i,L}, o_{i,L})\}$ and a final integrated response $f_i$, where $L$ denotes the total number of interactive steps. Specifically, each action $A_{i,t} = \langle n_{i,t}, a_{i,t}, p_{i,t} \rangle$ consists of a cognitive decision token (or thought) $n_{i,t}$, a selected tool $a_{i,t} \in \mathcal{T}$, and the formulated tool parameters $p_{i,t}$ grounded within the input $v_i$. Consequently, this cognitive orchestration task requires the model to dynamically schedule fine-grained, interrelated tools and process multi-step multimodal feedback, posing significantly greater challenges than static instruction following.

\begin{table}[h]
	\centering
	\caption{Taxonomy of tools in VAD-Tool. The tools are categorized based on the practical security governance pipeline: from entity tracking and behavior understanding to hazard detection.} 
	\label{tab: vad_tool_summary}
	\resizebox{\linewidth}{!}{ 
		\begin{tabular}{lll}
			\toprule
			\textbf{Primary Categories} & \textbf{Secondary Categories} & \textbf{Tool Inputs} \\
			\midrule
			\multirow{4}{*}{Entity Identity \& Tracking} 
			& Person Re-identification & Image, Video \\
			\cmidrule{2-3}
			& Gait Recognition & Video \\
			\cmidrule{2-3}
			& License Plate Recognition & Image, Video \\
			\cmidrule{2-3}
			& Car Re-identification & Image, Video \\
			\midrule
			\multirow{4}{*}{Behavior \& Scene Understanding} 
			& Pose Estimation & Video, Image \\
			\cmidrule{2-3}
			& Human Fall Detection & Video \\
			\cmidrule{2-3}
			& Crowd Counting & Image \\
			\cmidrule{2-3}
			& Scene Recognition & Video \\
			\midrule        
			\multirow{4}{*}{Security Event \& Hazard Detection} 
			& Violence Detection & Video \\
			\cmidrule{2-3}
			& Weapon Detection & Image, Text \\
			\cmidrule{2-3}
			& Fire and Smoke Detection & Video \\
			\cmidrule{2-3}
			& Anomaly Detection & Video, Text \\
			\bottomrule
		\end{tabular}
	}
\end{table}

\subsection{Fine-grained and Interrelated Tools}
To support complex multi-step reasoning, VAD-Tool encapsulates 12 distinct visual expert models as its action space $\mathcal{T}$, as shown in Table~\ref{tab: vad_tool_summary}. Inspired by the insights from \cite{zhong2025viotgpt}, the vast majority of public security incidents fundamentally revolve around tracing critical targets, interpreting abnormal behaviors, and identifying severe environmental hazards. Grounded in this practical taxonomy, we systematically organize our visual tools into three corresponding functional groups. This structured design comprehensively covers the essential dimensions of real-world video security analysis while rigorously testing the agent's fine-grained differentiation capabilities during interrelated events.

\noindent \textbf{Entity Identity \& Tracking:} This category targets the identification and continuous tracking of critical subjects and vehicles. It includes Person Re-identification (utilizing Fast-ReID \cite{He2023:FastReID} on Market-1501 \cite{Zheng2015:Market1501}), Gait Recognition (based on OpenGait \cite{Fan2023:OpenGait}, evaluated on CASIA-B \cite{Zheng2011:GaitViewTransform}), License Plate Recognition (evaluated using the Chinese City Parking Dataset, CCPD \cite{Xu2018:CCPD}), and Car Re-identification (evaluated on the VeRi dataset \cite{Liu2016:LargeScaleVehicleReID}).

\noindent \textbf{Behavior \& Scene Understanding:} Designed for dense spatial analysis and human action interpretation, this suite comprises Pose Estimation (RTMO \cite{lu2024rtmo}), Human Fall Detection (adapted from a YOLOv7-POSE-based model \cite{wang2023yolov7}), Crowd Counting (employing CLTR \cite{liang2022end} on the ShanghaiTech dataset \cite{zhang2016single}), and Scene Recognition (pre-trained Places365 \cite{Zhou2017:Places} mapped to UCF-Crime \cite{sultani2018real}).

\noindent \textbf{Security Event \& Hazard Detection:} This group focuses on identifying high-stakes, critical scenarios and secondary hazards. It encompasses Violence Detection (adapting contrastive vision-language models for fine-grained recognition \cite{wu2024vadclip}), a general Weapon Detection tool (powered by open-vocabulary grounding \cite{liu2024grounding}), Fire \& Smoke Detection (based on YOLOv8 \cite{varghese2024yolov8}), and an overarching Anomaly Detection expert \cite{zhang2025holmes} that leverages multi-modal large language models for comprehensive explanations.

To address the complexity of real-world VAD tasks,we explicitly design two interrelated scenarios to evaluate multi-tool synergy: Pose-based Violence Detection and Scene Anomaly Detection. Specifically, Pose-based Violence Detection integrates human pose estimation with violence recognition, whereas Scene Anomaly Detection combines spatial scene understanding with general anomaly identification.

\subsection{Human-in-the-Loop Data Annotation}
The instruction data and reasoning trajectories are collected via a rigorous, four-step human-in-the-loop (HITL) pipeline: Defining Visual Tool Categories, Collecting Raw Data, Generating ReAct Annotations, and Expert Review. To ensure reasoning practicability, we utilize strong vision-language models (specifically Qwen2.5-VL) to formulate the initial multi-step ground-truth trajectories into the standard ReAct format \cite{Yao2023:ReAct}, encompassing step-by-step Thought, Action, Action Input, and Observation. Subsequently, Large Language Models (\emph{e.g.}, Qwen2.5-VL) are employed to semantically augment the human linguistic queries $q_i$. This ensures extreme linguistic diversity in user inputs (as evidenced by our diverse word cloud of expressions), while strictly preserving the underlying causal tool-call trajectories. Finally, domain experts thoroughly review the generated pairs as the crucial "human-in-the-loop" to ensure zero-tolerance logical correctness. 

\begin{figure}[t]
    \centering
    \includegraphics[width=0.99\columnwidth]{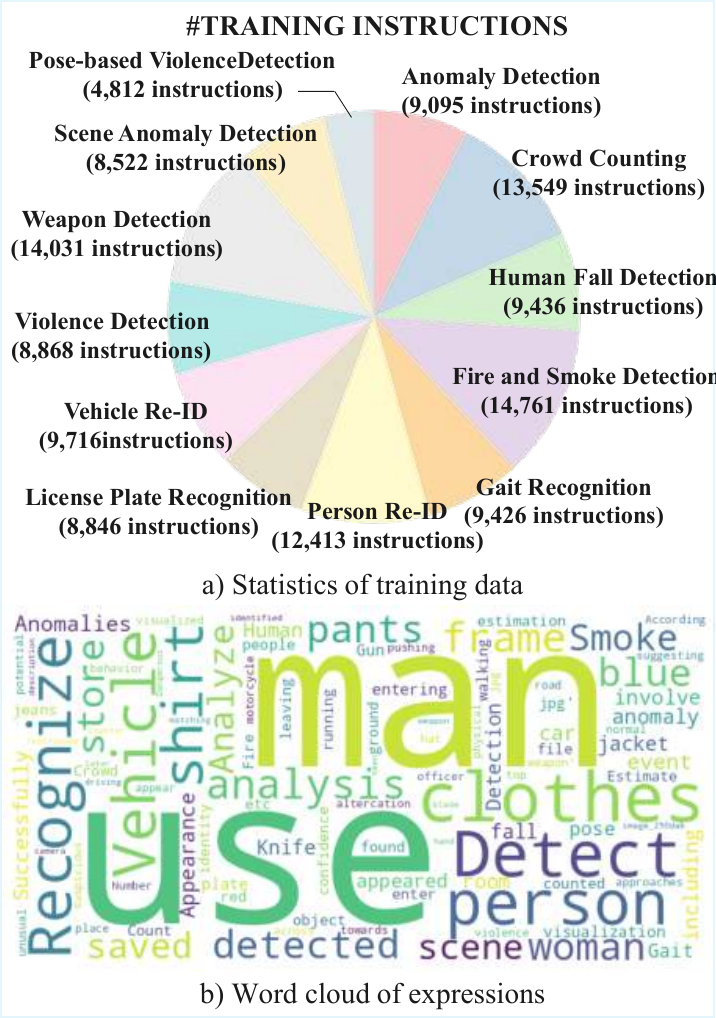}
   \caption{Overview of VAD-Tool dataset construction and statistics. (a) Training data distribution across tool categories. (b) Word cloud of instruction queries. }
    \label{fig: dataset statistics}
\end{figure}

\subsection{Dataset Statistics}
Through meticulous annotation and augmentation, the VAD-Tool benchmark comprises a large-scale training dataset containing approximately 15.48 million tokens. As illustrated in our data statistics (Figure~\ref{fig: dataset statistics}), the dataset features a well-balanced and highly diverse distribution of training instructions across various high-stakes scenarios. Specific categories include Fire and Smoke Detection ($\sim$14.8k instructions), Weapon Detection ($\sim$14.0k), Crowd Counting ($\sim$13.5k), Person Re-ID ($\sim$12.4k), Vehicle Re-ID ($\sim$9.7k), Human Fall Detection ($\sim$9.4k), Gait Recognition ($\sim$9.4k), general Anomaly Detection ($\sim$9.1k), Violence Detection ($\sim$8.9k), License Plate Recognition ($\sim$8.8k), Scene Anomaly Detection ($\sim$8.5k), and fine-grained Pose-based Violence Detection ($\sim$4.8k), cumulatively scaling to over 120k instruction-trajectory pairs.

For rigorous evaluation, we construct a corresponding testing dataset of 1,841 multi-step reasoning samples. To prevent data leakage and evaluate true cognitive generalization, the physical video scenes and the linguistic semantics in the testing benchmark are strictly distinct from those in the training dataset, while maintaining a consistent overall tool-call distribution to eliminate evaluation bias.

\begin{figure*}[ht]
	\centering
	\includegraphics[width=0.99\textwidth]{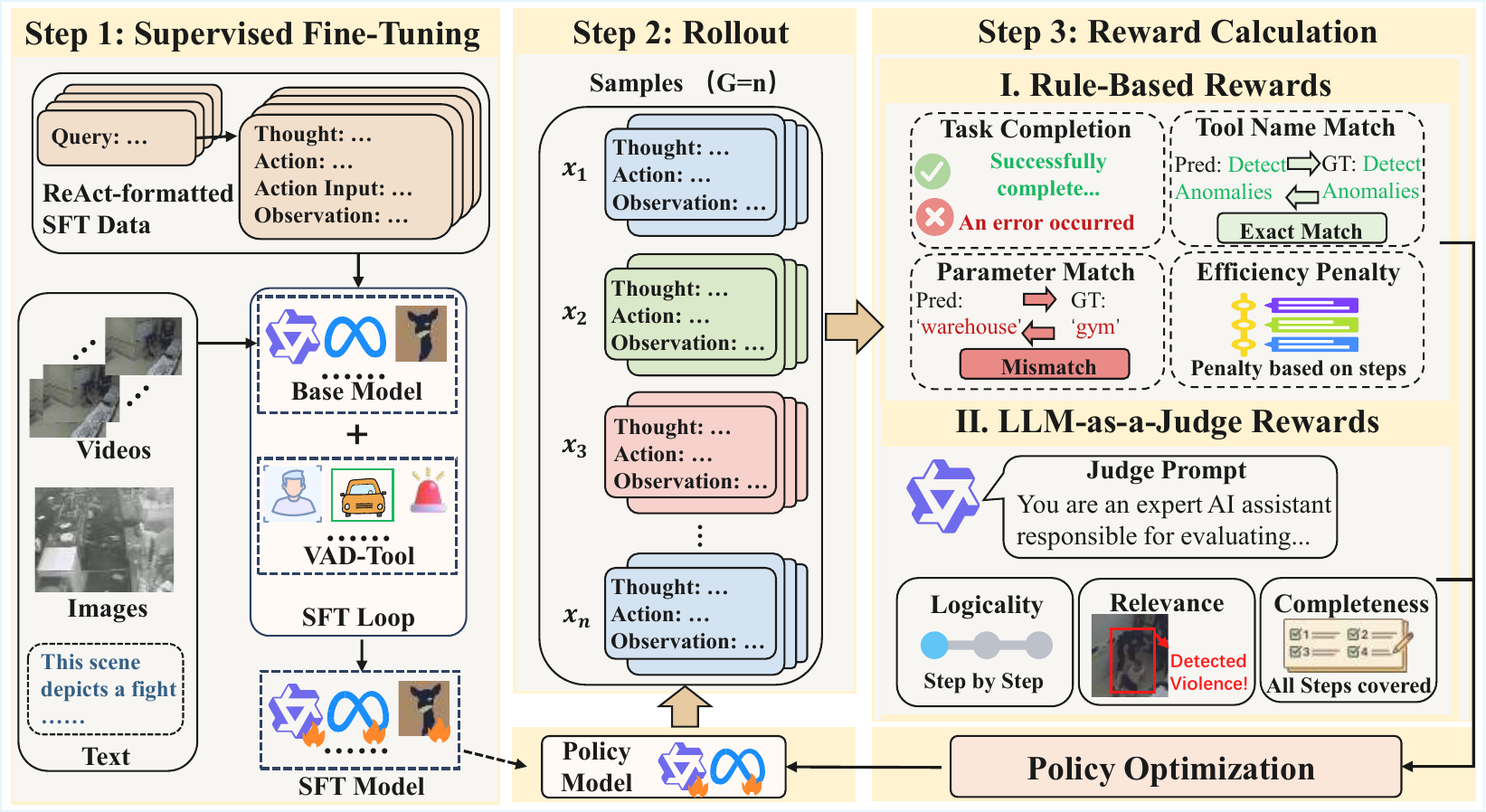}
	\caption{The overall pipeline of the VTO framework. Step 1: Supervised Fine-Tuning  equips the base model with foundational tool-use capabilities via ReAct-formatted data. To achieve process-supervised cognitive alignment, the agent enters a continuous learning loop. In Step 2 (Rollout), the policy model explores multiple reasoning trajectories in parallel. In Step 3 (Reward Calculation), these trajectories are dynamically evaluated by synergizing Rule-Based  Rewards and LLM-as-a-Judge  Rewards, providing comprehensive cognitive supervision for final Policy Optimization.}
    \label{fig: VTO}
\end{figure*}

\section{VTO Framework}

\subsection{Overview}

As illustrated in Figure~\ref{fig: VTO}, the VTO framework operates in two stages: Supervised Fine-Tuning (SFT) and Process-Supervised Cognitive Alignment. While Step 1 (SFT) establishes the foundational ReAct formatting for basic tool invocation, it inherently relies on static behavior cloning. This static paradigm suffers from exposure bias and struggles to maintain the strict logical interdependencies and multi-step causal constraints required in complex VAD scenarios. To overcome this, Step 2 formulates the interrelated tool scheduling as a reinforcement learning problem optimized via Group Relative Policy Optimization (GRPO) \cite{shao2024deepseekmath}. During the rollout phase, the agent explores diverse reasoning trajectories by dynamically interacting with the visual environment. To address the unique zero-tolerance safety requirements of security events, we design a Process-Supervised Cognitive Alignment method that jointly leverages deterministic rule-based metrics and LLM-as-a-Judge evaluations. Rather than treating RL as a generic black box, this dual-reward framework is explicitly tailored to penalize premature termination and hallucinated tool routing, thereby guiding the policy to converge on robust and interrelated causal chains.

\subsection{Supervised Fine-Tuning (SFT)}
To bridge the gap between general-purpose base models and specialized security agents, we first perform Supervised Fine-Tuning (SFT) on our meticulously annotated VAD-Tool dataset. Beyond mere behavior cloning, this stage is essential to endow the policy model with three critical domain-specific capabilities: 
(1) \textbf{Syntactic Tool Invocation}: Conditioning the agent to strictly adhere to valid, parsable ReAct formatting for complex visual tool scheduling.
(2) \textbf{Anomaly Event Modeling}: Adapting the model's internal representations to understand and reason over complex physical anomalies and high-stakes security events, which are severely underrepresented in general pre-training corpora.
(3) \textbf{Observation Integration}: Learning to comprehend and contextualize the highly specific, and potentially unfamiliar, observations returned by these visual experts (\emph{e.g.}, tracking IDs, spatial coordinates, or raw pose keypoints).

The SFT process formulates multi-step trajectory generation as a standard auto-regressive next-token prediction task. Specifically, the policy $\pi_\theta$ is optimized by minimizing the standard negative log-likelihood loss:
\begin{equation}
     \mathcal{L}_{\mathrm{sft}} = -\sum_{i}\sum_{t} \log P_{\theta}(A_{i,t} | q_i, v_i, C_{i,t-1}),
\end{equation}

\noindent where $i$ and $t$ denote the indices of the training sample and the current reasoning step, respectively. $P_{\theta}$ represents the probability distribution predicted by the model parameterized with $\theta$. At each step $t$, the model is trained to generate the target action $A_{i,t}$ (comprising the cognitive thought, selected tool, and formulated parameters) conditioned on the natural language query $q_i$, the physical visual environment $v_i$, and the accumulated historical reasoning trajectory $C_{i,t-1}$.

\subsection{Process-Supervised Cognitive Alignment}
\label{sec:rl_framework}

While Supervised Fine-Tuning (SFT) provides a necessary initialization for tool usage, it fundamentally relies on static behavior cloning. This static paradigm struggles to generalize to the complex logical interdependencies and stringent safety constraints inherent in VAD scenarios. To break this performance upper bound and align the agent's reasoning process with expert-level logical rigor, we transition the training paradigm to a dynamic, process-supervised cognitive alignment method.

As illustrated in Figure~\ref{fig: VTO}, our proposed VTO framework establishes a continuous alignment loop comprising three primary phases: Rollout, Reward Calculation, and Policy Optimization.

\noindent \textbf{Rollout.} In this phase, the policy model actively explores the action space by sampling multiple diverse reasoning trajectories (denoted as $x_1, \dots, x_n$) for a given visual query. Instead of following a single fixed ground-truth path, the agent iteratively generates cognitive Thoughts, selects Actions, formulates Action Inputs, and receives dynamic Observations from the VAD-Tool environment. This continuous interaction allows the model to accumulate rich, multi-step experiences for subsequent evaluation.

The detailed mathematical formulations of the Process-Supervised Cognitive Alignment and the GRPO objective are elaborated subsequently in Section~\ref{sec:prm} and Section~\ref{sec:grpo}.



    

\subsubsection{Process-Supervised Cognitive Alignment}
\label{sec:prm}

While Supervised Fine-Tuning equips the agent with basic tool-invocation abilities, it suffers from exposure bias and fails to explicitly penalize critical logical errors (\emph{e.g.}, premature termination) in multi-step scenarios. To provide dense, step-wise cognitive supervision, we propose a comprehensive Process-Supervised Cognitive Alignment method. As illustrated in Step 3 of Figure~\ref{fig: VTO}, this method evaluates the sampled trajectories using a dual-reward system: Rule-Based Objective Rewards and LLM-as-a-Judge  Rewards.

Formally, at each reasoning step $t$ (or upon trajectory completion), the overall reward $r_t$ is a combination of  rules and LLM evaluations:
\begin{equation}
    r_t =  \mathcal{R}_{\text{rule}} +  \mathcal{R}_{\text{LLM}}.
\end{equation}

\textbf{I. Rule-Based Rewards ($\mathcal{R}_{\text{rule}}$):} This component provides strict, deterministic feedback based on ground-truth alignments and execution efficiency. It is defined as:
\begin{equation}
    \mathcal{R}_{\text{rule}} = \mathcal{R}_{\text{comp}} + \mathcal{R}_{\text{tool}} + \mathcal{R}_{\text{param}} + \mathcal{R}_{\text{eff}}.
\end{equation}
Here, $\mathcal{R}_{\text{comp}}$ assigns positive scores for successful task completion and negative penalties for execution errors. $\mathcal{R}_{\text{tool}}$ and $\mathcal{R}_{\text{param}}$ enforce exact matches between the predicted tool names/parameters and the ground truth. Finally, $\mathcal{R}_{\text{eff}}$ applies an efficiency penalty proportional to the number of steps taken, explicitly discouraging redundant tool invocations.

\textbf{II. LLM-as-a-Judge Rewards ($\mathcal{R}_{\text{LLM}}$):} To assess the qualitative aspects of the reasoning chain that rigid rules might miss, we employ Qwen2.5-VL-72B, a 72B-parameter model selected for its instruction-following and reasoning capabilities, as the expert evaluator. Guided by a specific "Judge Prompt", it scores the trajectory based on three cognitive dimensions:
\begin{equation}
    \mathcal{R}_{\text{LLM}} = \mathcal{R}_{\text{logic}} + \mathcal{R}_{\text{rel}} + \mathcal{R}_{\text{complete}}.
\end{equation}
Specifically, Logicality ($\mathcal{R}_{\text{logic}}$) evaluates the step-by-step causal reasoning; Relevance ($\mathcal{R}_{\text{rel}}$) measures how well the chosen actions and visual observations align with the user's initial query; and Completeness ($\mathcal{R}_{\text{complete}}$) ensures that all necessary logical steps are fully covered before generating the final response, thereby penalizing premature termination.

This dual-reward structure ensures that early logical mistakes and inefficient explorations are immediately penalized, preventing the agent from exploiting flawed trajectories. The cumulative return for a trajectory of length $L$ is given by $R(\tau) = \sum_{t=1}^{L} \gamma^{t-1} r_t$, where $\gamma$ is the discount factor.

\begin{table*}[!h]
\caption{Quantitative results on VAD-Tool. "Decis" represents the accuracy of decisions of whether to use tools $Acc_{n_{i,t}}$. "Tool" represents the accuracy of chosen tool names $Acc_{a_{i,t}}$. "Input" represents the accuracy of input information of tools $Acc_{p_{i,t}}$. "Whole" represents the accuracy of the whole response $Acc_{A_{i,t}}$. Best results are in \textbf{bold} and the second best are \underline{underlined}. }
\label{tab:main_results}
\centering
\resizebox{\textwidth}{!}{%
\begin{tabular}{@{}l|l|cccc|cccc|cccc@{}}
\toprule
\multirow{2}{*}{\textbf{Models}} & \multirow{2}{*}{\textbf{Prompt}} & \multicolumn{4}{c|}{\textbf{Single Tool Responses}} & \multicolumn{4}{c|}{\textbf{Interrelated Tools Responses}} & \multicolumn{4}{c}{\textbf{All Responses}} \\ \cmidrule(l){3-14} 
 & & \textbf{Decis(\%)} & \textbf{Tool(\%)} & \textbf{Input(\%)} & \textbf{Whole(\%)} & \textbf{Decis(\%)} & \textbf{Tool(\%)} & \textbf{Input(\%)} & \textbf{Whole(\%)} & \textbf{Decis(\%)} & \textbf{Tool(\%)} & \textbf{Input(\%)} & \textbf{Whole(\%)} \\ \midrule
Vicuna-7B & Zero-shot & 74.52 & 54.03 & 22.93 & 20.66 & 0 & 0 & 0 & 0 & 66.43 & 48.16 & 20.44 & 18.42 \\
Vicuna-7B & Few-shot &18.85 &1.62 &0 &0 & 0 & 0 & 0 & 0 & 16.80 & 1.44 & 0.00 & 0.00 \\
Vicuna-7B (SFT) & Zero-shot &75.00 & 55.65& 40.63&39.84 & 0 & 0 & 0 & 0 & 66.86 & 49.61 & 36.22 & 35.51 \\ 
\midrule
Llama-3-8B & Zero-shot & 28.78 & 29.08 & 24.33 & 24.33 & 16.58& 18.59&5.53 &4.52 & 27.46 & 27.94 & 22.29 & 22.18 \\
Llama-3-8B & Few-shot &24.49 &22.80 & 20.66& 20.53&69.59 &68.13 & 43.22& 42.85& 29.39 & 27.72 & 23.11 & 22.95 \\
Llama-3-8B (SFT) & Zero-shot & 95.45 & 95.45 & 82.91 & 82.90 & 92.67 & 92.67 & 65.93 & 65.93 & 95.15 & 95.15 & 81.07 & 81.06 \\
\midrule
Qwen3-VL-8B & Zero-shot & 98.03 & 89.96 & 77.68 & 77.41 & 80.58 & 80.58 & 64.46 & 63.73 & 96.14 & 88.94 & 76.24 & 75.92 \\
Qwen3-VL-8B & Few-shot & \textbf{100.00} & 91.92&74.43 &74.43 &89.01 &89.01 &63.73 &63.00 & 98.81 & 91.60 & 73.27 & 73.19 \\
Qwen3-VL-8B (SFT) & Zero-shot & 98.97 & 98.84 & 85.76 & 85.76 & 99.26 & 99.26 & \underline{90.47} & \underline{89.74} & 99.00 & 98.89 & \underline{86.27} & \underline{86.19} \\ \midrule
Qwen2.5-VL-72B & Zero-shot &96.42 & 98.58&81.86 & 81.49&88.64 & 97.80&76.58 & 67.40& 95.58 & 98.50 & 81.29 & 79.96 \\
Qwen2.5-VL-72B & Few-shot & \underline{99.06}& 90.45&72.58 &72.43 &80.95 & 93.41&78.02 &66.30 & 97.09 & 90.77 & 73.17 & 71.76 \\ \midrule
VTO (Llama-3-8B) & Zero-shot & 99.02 & \underline{99.02} & \underline{87.82} & \underline{87.82} & \textbf{100.00} & \textbf{100.00} & 73.26& 72.16& \underline{99.13} & \underline{99.13} & 86.24 & 86.12 \\
VTO (Qwen3-VL-8B) & Zero-shot & \textbf{100.00} & \textbf{100.00} & \textbf{96.54} & \textbf{96.54} & \textbf{100.00} & \textbf{100.00} & \textbf{95.60} & \textbf{95.14} & \textbf{100.00} & \textbf{100.00} & \textbf{96.44} & \textbf{96.39} \\ \bottomrule
\end{tabular}
}
\end{table*}

\subsubsection{Policy Optimization via GRPO}
\label{sec:grpo}

Driven by the comprehensive reward signals from the process reward mechanism (PRM), we optimize the LLM agent $\pi_\theta$ using Group Relative Policy Optimization (GRPO). Unlike standard PPO that requires a memory-intensive value network, GRPO significantly reduces training overhead by estimating the baseline directly from a group of sampled trajectories. This memory efficiency is particularly crucial in our multimodal VAD domain, where processing long visual contexts and accumulating multi-step tool observations inherently consumes substantial computational resources.

During the rollout phase, for each human query $q_i$ and visual input $v_i$, the agent $\pi_\theta$ samples a group of $G$ distinct reasoning trajectories $\{C_{i,L}^g\}_{g=1}^G$ according to the policy $\pi_\theta(\cdot | q_i, v_i, \mathcal{T})$. For each trajectory $C_{i,L}^g$, we calculate its cumulative process reward $R_g = \sum_{t=1}^{L_g} \gamma^{t-1} r_t$ using the PRM evaluations. Instead of relying on a value network to estimate the baseline, GRPO computes the advantage $\hat{A}_g$ by normalizing the rewards within the group:
\begin{equation}
    \hat{A}_g = \frac{R_g - \text{mean}(\{R_1, \dots, R_G\})}{\text{std}(\{R_1, \dots, R_G\}) + \epsilon_{std}}.
\end{equation}
This group-relative advantage effectively indicates whether a specific tool-scheduling trajectory performs better or worse than the average exploration attempt for that specific query.

To ensure stable policy updates and prevent catastrophic degradation during exploration, we maximize the clipped surrogate objective combined with a Kullback-Leibler (KL) divergence penalty. The objective function for GRPO is formulated as:
\begin{multline}\label{eq:grpo}
\mathcal{L}^{\mathrm{GRPO}}(\theta) = \hat{\mathbb{E}}_{q \sim \mathcal{Q}, \tau \sim \pi_{\theta_{old}}} \Bigg[ \frac{1}{G} \sum_{g=1}^{G} \sum_{t=1}^{L_g} \min \left( \rho_t(\theta) \hat{A}_g, \right. \\
\left. \text{clip}(\rho_t(\theta), 1-\epsilon, 1+\epsilon) \hat{A}_g \right) - \beta \mathbb{D}_{KL}(\pi_\theta \| \pi_{ref}) \Bigg],
\end{multline}
where $\rho_t(\theta) = \frac{\pi_\theta(A_{i,t} | C_{i,t-1})}{\pi_{\theta_{old}}(A_{i,t} | C_{i,t-1})}$ is the probability ratio between the current policy and the old policy, $\epsilon$ is the clipping hyperparameter. $\pi_{ref}$ is the reference model (typically the initial SFT model), and $\beta$ is the KL penalty coefficient used to constrain the policy drift.

By iteratively updating $\theta$ via Equation~(\ref{eq:grpo}), VTO breaks the performance upper bound of static behavior cloning without the massive memory footprint of a critic model. The dual-reward process-supervised reinforcement learning mechanism forces the LLM to strictly align its internal reasoning states with the complex logical interdependencies required by intelligent Social Security Governance.

\section{Experiments}
\subsection{Experimental Settings}
\noindent \textbf{Baselines.}
With limited computing
resources, we mainly investigate how to leverage LLMs with tens of billions of parameters effectively. We use Llama-3-8B~\cite{Touvron2023:LLaMA}, Vicuna-7B~\cite{chiang2023vicuna}, and Qwen3-VL-8B~\cite{bai2025qwen3} as base models for the following fine-tuning. Correspondingly, Llama-3-8B, Vicuna-7B, and Qwen3-VL-8B without fine-tuning are used as baselines, which rely on the in-context ability with the same prompt. Qwen2.5-VL-72B is also included as a strong large-scale baseline for comparison on VAD-Tool.

\noindent\textbf{Evaluation Metrics.} We employ four metrics to evaluate the agent's performance. For the intermediate reasoning tuple $A_{i,t} = \langle n_{i,t}, a_{i,t}, p_{i,t} \rangle$ at step $t$ of query $i$, let $\hat{A}_{i,t} = \langle \hat{n}_{i,t}, \hat{a}_{i,t}, \hat{p}_{i,t} \rangle$ denote the corresponding prediction. The tuple components represent the decision of whether to use tools (\emph{e.g.}, "Thought: Do I need to use a tool? Yes/No") $n_{i,t}$, the tool selection $a_{i,t}$, and the input parameter $p_{i,t}$. We evaluate the exact match accuracy for these three step-level metrics: decision accuracy (\textbf{Decis}, $Acc_{n_{i,t}}$), tool selection accuracy (\textbf{Tool}, $Acc_{a_{i,t}}$), and input parameter accuracy (\textbf{Input}, $Acc_{p_{i,t}}$). For any given component $x \in \{n, a, p\}$, its accuracy is formally unified as:
\begin{equation} \label{eq:acc_step}
Acc_{x_{i,t}} = \frac{1}{N} \sum_{i=1}^{N} \mathbb{I} \big( \forall t \in \{1, \dots, T_i\}, \hat{x}_{i,t} = x_{i,t} \big),
\end{equation}
where $N$ is the total number of evaluation queries, $T_i$ is the total number of reasoning steps for query $i$, and $\mathbb{I}(\cdot)$ is the indicator function.


\noindent \textbf{Implementation Details.} Our models are trained on NVIDIA RTX 4090 GPUs via a two-stage pipeline. Initially, we perform Supervised Fine-Tuning (SFT) for 1 epoch using parameter-efficient LoRA ($r=8, \alpha=16, \text{dropout}=0.05$) with a learning rate of 5e-5 and a batch size of 64. Subsequently, for the Cognitive Alignment stage via GRPO, we transition to full-parameter tuning for 5 epochs. During this RL phase, we employ SGLang for asynchronous rollout ($G=8$), applying a batch size of 16, a strictly lower learning rate of 1e-6, and a KL penalty $\beta=0.001$ with maximum prompt and response lengths capped at 2048 and 1024 tokens respectively. The overall framework yields an average inference time of approximately 30 seconds per sample.

\subsection{Quantitative Results}
\label{sec:main_results}

\noindent \textbf{Response Results.} We first evaluate the fundamental capability of different paradigms in orchestrating visual tools under strict physical constraints. As reported in Table~\ref{tab:main_results}, while early baselines achieve marginal success on isolated single-tool tasks, they experience catastrophic performance degradation when confronted with Interrelated Tools. For instance, the interrelated whole-response accuracy of Vicuna-7B plummets to $0\%$, and Llama-3-8B (Zero-shot) drops to a mere 4.52\%. Even a massive model like Qwen2.5-VL-72B hits a upper bound around 67.40\% on interrelated tasks. Although standard supervised fine-tuning (SFT) partially mitigates format errors and significantly improves the $Decis$ and $Tool$ selection metrics, it still hits a severe bottleneck. For example, while Qwen3-VL-8B (SFT) performs well on single tools, it plateaus at 89.74\% in the interrelated $Whole$ metric. This indicates that purely mimicking trajectories via static behavior cloning is insufficient for complex spatial-temporal orchestration. In contrast, our proposed VTO framework demonstrates a decisive breakthrough. By formulating tool scheduling as a process-supervised cognitive alignment, VTO significantly outperforms both zero-shot baselines and SFT models. Specifically, on the Qwen3-VL-8B backbone, this paradigm pushes the interrelated $Whole$ accuracy to an impressive 95.14\% (achieving a perfect 100\% in Decis and Tool selection). Similarly, VTO based on Llama-3-8B improves the interrelated performance from 65.93\% to 72.16\%. By leveraging the Process Reward Mechanism, VTO effectively reduces critical tool-orchestration errors and better aligns the agent's behavior with the stringent safety requirements of real-world surveillance scenarios.

\begin{figure*}[ht]
	\centering
    \includegraphics[width=0.99\textwidth]{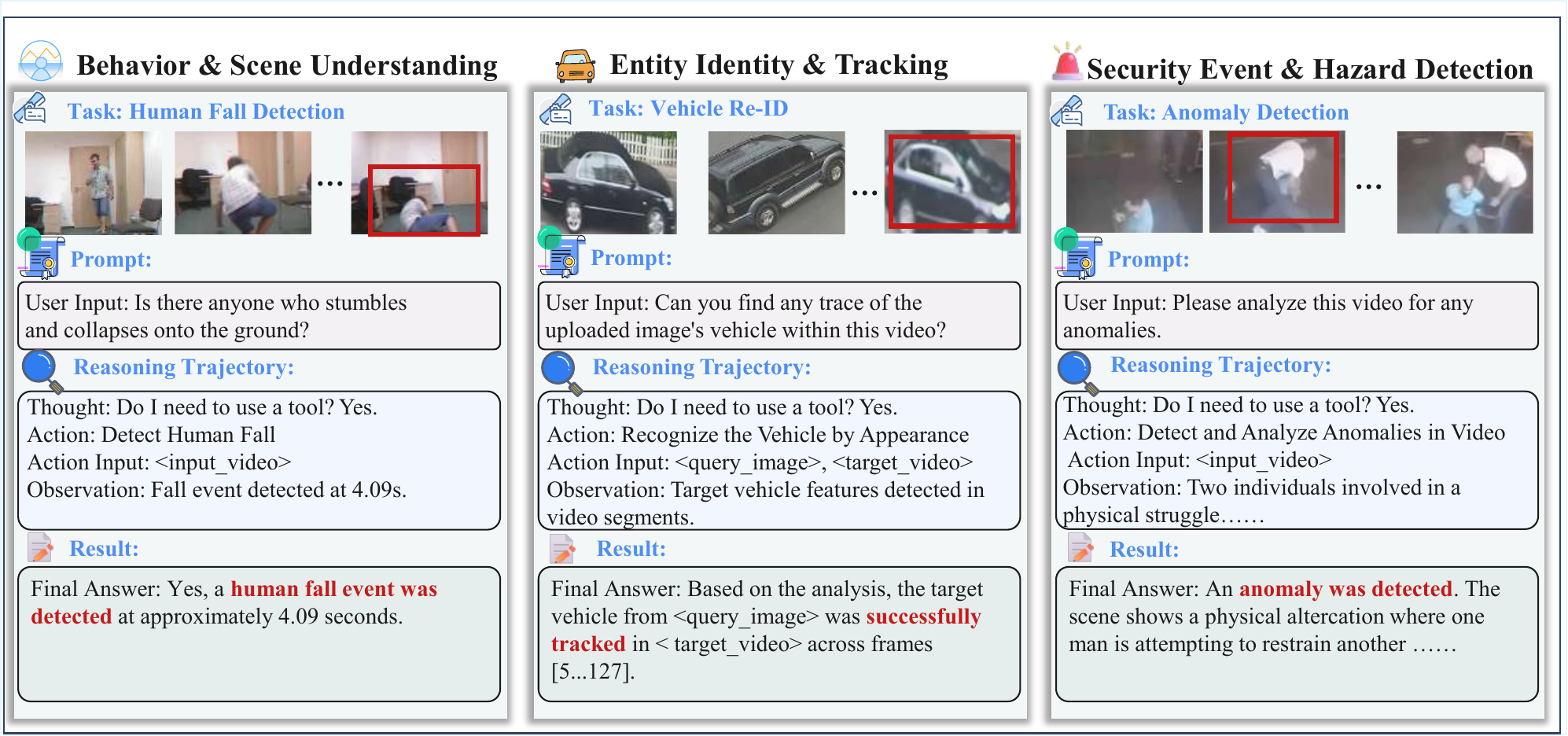}
    \caption{Qualitative examples from the VAD-Tool benchmark illustrating structured reasoning trajectories across three diverse domains: Behavior \& Scene Understanding, Entity Identity \& Tracking, and Security Event \& Hazard Detection. Notably, the red highlighted text in the final answers directly corresponds to the red bounding boxes in the visual frames.}
    \label{fig: tool usage}
\end{figure*}

\begin{table}[ht]
\centering
\caption{Ablation study on the Process-Supervised Cognitive Alignment using Qwen3-VL-8B. The evaluation is conducted on Interrelated Tools Responses to highlight the impact of different reward granularity. Best results are in \textbf{bold} and the second best are \underline{underlined}.}

\resizebox{\columnwidth}{!}{%
\begin{tabular}{@{}l|cccc@{}}
\toprule
\multirow{2}{*}{\textbf{Paradigm}} & \multicolumn{4}{c}{\textbf{Interrelated Tools Responses}} \\ \cmidrule(l){2-5} 
 & \textbf{Decis(\%)} & \textbf{Tool(\%)} & \textbf{Input(\%)} & \textbf{Whole(\%)} \\ \midrule
Baseline (SFT Only) &  \underline{99.26} & \underline{99.26} & 90.47 & 89.74  \\ \midrule
\textbf{Rule-Based Rewards} & & & & \\
\quad w/o Task Completion Reward & \textbf{100.00} & \textbf{100.00}  &29.92  & 29.92 \\
\quad w/o Tool Name Match Reward & \textbf{100.00} & 94.66 & \textbf{98.04} & 93.01  \\
\quad w/o Tool Param Match Reward & \textbf{100.00} & \textbf{100.00} & 78.02 & 77.66 \\
\quad w/o Efficiency Reward & \textbf{100.00} & \textbf{100.00} &94.48 & \underline{94.48}  \\ \midrule
\textbf{LLM-as-a-Judge  Reward} & & & & \\
\quad w/o LLM Judge Reward & \textbf{100.00} & \textbf{100.00} & 94.97 & 93.70  \\ \midrule
\textbf{VTO (Ours)} & \textbf{100.00} & \textbf{100.00} & \underline{95.60} & \textbf{95.14} \\ \bottomrule
\end{tabular}%
}
\label{tab:ablation_reward}
\end{table}

\noindent \textbf{Ablation Study.} 
To verify the indispensability of our Process-Supervised Cognitive Alignment, we conduct a rigorous ablation study on Qwen3-VL-8B (Table~\ref{tab:ablation_reward}). Compared to the SFT baseline (89.74\%), our \textbf{VTO} achieves a substantial performance leap in overall multi-step execution accuracy (95.14\%). Decomposing VTO reveals that fine-grained objective anchors are critical for environmental grounding. Notably, removing the $\mathcal{R}_{\text{comp}}$ causes a catastrophic performance collapse (29.92\%), indicating that a strict global signal is essential to prevent the agent from trapping in endless tool-use loops. Similarly, without the $\mathcal{R}_{\text{param}}$, input accuracy drops sharply to 78.02\%, proving explicit parameter supervision prevents hallucinatory video frame or spatial coordinate grounding. Finally, while rule-based metrics ensure syntactic correctness, ablating the $\mathcal{R}_{\text{LLM}}$ also degrades overall performance (93.70\%). This confirms the LLM evaluator's important role in maintaining the logical relevance and causal completeness of the reasoning chains.

\subsection{Qualitative Results}
\label{sec:case_study}


\noindent \textbf{VAD-Tool Benchmark Demonstration.}~Figure~\ref{fig: tool usage} illustrates the VAD-Tool benchmark's structured reasoning trajectories across diverse domains: \textbf{Human-centric}, \textbf{Vehicle \& Traffic}, and \textbf{Environment \& Hazard Analysis}. Each category displays a full reasoning cycle from multimodal user query to final integrated linguistic response. These examples highlight the benchmark's capability to validate an agent's fundamental grounding and single-step tool interaction skills in unconstrained physical environments, establishing a robust foundation.


\section{Conclusion}
In this paper, we presented VTO, a dynamic multimodal agent powered by process-supervised reinforcement learning for explainable Video Anomaly Detection (VAD), alongside VAD-Tool, a comprehensive benchmark for interrelated anomaly reasoning. To overcome the imitation-based limitations of Supervised Fine-Tuning (SFT) and explicitly address the premature termination problem inherent in standard outcome-based RL, VTO formulated visual tool scheduling as a dynamic policy optimization process. Our framework provided fine-grained, step-wise cognitive supervision across logical thoughts, tool selections, and parameter formulations.

\begin{acks}
This work is partly supported by the Funds for the National Natural Science Foundation of China under Grant 62572072 and Beijing Natural Science Foundation (L243027).
\end{acks}

\bibliographystyle{ACM-Reference-Format}
\bibliography{sample-base}

\appendix

\end{document}